# Transforming Color: Novel Image Colorization Method

**Hamza Shafiq, and Bumshik Lee***

Hamza Shafiq and Bumshik Lee are affiliated with Chosun University, Department of ICE, 309 Pilmundaero, Gwangju, South Korea, 61452.

*Correspondence: bslee@chosun.ac.kr

## Abstract

This paper introduces a novel method for image colorization that utilizes a Color Transformer and Generative Adversarial Networks (GANs) to address the challenge of generating visually appealing colorized images. Conventional approaches often struggle with capturing long-range dependencies and producing realistic colorizations. The proposed method integrates a transformer architecture to capture global information and a GAN framework to improve visual quality. In this study, a Color Encoder that utilizes a random normal distribution to generate color features is applied. These features are then integrated with grayscale image features to enhance the overall representation of the images. Our method demonstrates superior performance compared with existing approaches by utilizing the capacity of the transformer, which can capture long-range dependencies and generate a realistic colorization of GAN. Experimental results show that the proposed network significantly outperforms other state-of-the-art colorization techniques, highlighting its potential for image colorization. This research opens new possibilities for precise and visually compelling image colorization in domains such as digital restoration and historical image analysis.

## Keywords

Image Colorization; Transformer; Generative Adversarial Network.

## 1. Introduction

Adding colors to grayscale or black-and-white images is known as image colorization. This technology holds considerable importance across multiple fields, such as the digital restoration of old photographs, entertainment and media sectors, historical preservation, and augmentation of visual communication. Incorporating color into images can enhance their realism and visual appeal by accurately representing the depicted scene or object.

Despite its significance, image colorization poses numerous challenges. One of the foremost challenges is the precise selection of suitable colors for individual pixels within an image, particularly when color data are unavailable. The difficulty level of this task escalates when confronted with complex visuals or uncertain grayscale variations. Throughout the years, researchers have proposed various methodologies to address the issue of image colorization. Historically, conventional techniques have frequently required human involvement, whereby skilled artists or experts have meticulously incorporated hues into monochromatic images. Although these methods produced adequate outcomes, they were time-intensive and required knowledge of color theory and image manipulation.

Automated image colorization techniques have attracted significant attention recently owing to the progress made in deep learning and computer vision. These methodologies utilize extensive datasets, convolutional neural networks (CNNs), and generative models [1] to learn the correlation between grayscale and color images. In this study, we aim to develop an automated system for predicting and assigning appropriate colors to grayscale pixels, thereby significantly reducing the need for manual intervention.

Although current methods leverage CNNs or transformer architectures, they face challenges, such as color bleeding, desaturation, and limitations in effectively capturing local and global features. The trickier part of colorization is determining the right balance between paying attention to small details, such as textures and comprehending a broader context. These challenges make it difficult for automated methods to consistently produce accurate and visually pleasing colorizations. Striking a delicate balance between preserving fine details and comprehending a broader context is crucial for achieving natural and realistic results. Overcoming these challenges is essential for advancing state-of-the-art image colorization and ensuring that automated techniques seamlessly capture fine details for colorization.

This paper presents a new method for image colorization that overcomes certain limitations of current methodologies. The proposed method utilizes a color encoder [2], a color transformer, and encoder-decoder-based generative adversarial network (GAN) [1] architecture to achieve precise and effective colorization. The objective of incorporating a color transformer and encoder into the generator architecture is to improve the colorization procedure by providing color assignments that are more contextually relevant and coherent.

The subsequent sections of this paper are organized as follows. Section 2 gives a detail review of existing works, providing insight into existing methodologies and their limi-tations. Following this, Section 3 elaborates on our proposed method, detailing its theo-retical basis and practical implementation. Subsequently, Section 4 presents the results of our experimental evaluation, which includes comparisons with other state-of-the-art techniques. Finally, Section 5 concludes the paper, summarizing key findings and out-lining directions for future research.

## 2. Related Works

This section provides a comprehensive overview of the current image colorization methodologies, including conventional and deep-learning-based techniques. The strengths, limitations, and areas for improvement of the subject are analyzed, and highlights the specific gaps in existing research that our proposed method addresses.

Historically, conventional methods of colorizing images have relied heavily on the involvement of skilled artists and professionals. The aforementioned techniques entail a rigorous procedure for incorporating hues into monochromatic images based on color theory and artistic proficiency. An example of such a methodology is the research conducted by Levin *et al*., who presented a colorization technique based on scribbles [3]. Although manual techniques can produce acceptable outcomes, they are time-consuming, labor-intensive, and require proficient human operators. Traditional example-based methods were crucial in early attempts at image colorization. Approaches such as optimization techniques using graph cuts, energy minimization, [4] texture-based methods involving texture synthesis, and patch-based techniques [5] have been explored. These methods often rely on transferring color information from reference or exemplar images to grayscale targets, although they can face challenges in handling complex scenes and may introduce artifacts, such as color bleeding. The manual selection of reference images is also time-consuming.

Automated image colorization techniques have emerged as a promising approach owing to advancements in deep learning and computer vision. Deep learning techniques utilize extensive datasets and convolutional neural networks (CNNs) to understand the complex associations between grayscale and color images comprehensively. One noteworthy technique employs a deep learning-based strategy that incorporates both classification and colorization networks [6].

GANs have recently gained considerable attention. Using generative models enables multimodal colorization. In a recent study [1], a conditional GAN-based image-to-image translation model was proposed utilizing a generator based on the U-Net architecture. The results of this approach demonstrate improved image colorization, which can be attributed to the use of adversarial training. In [7], the model was extended to include high-resolution images. The generative priors for colorization was further investigated in [8], provided that the spatial structures of the image had already been produced. SCGAN [9] is a GAN-based image colorization method that uses saliency maps to guide the colorization. SCGAN can first focus on the most significant portions of an image by employing saliency maps, resulting in more accurate and realistic colorization results. The double-channel-guided GAN (DCGAN) [10] is another GAN-based image colorization approach and guides the colorization process using two channels, where the first channel contains a grayscale image, and the second includes a color palette. DCGAN learns the structure of an image using grayscale images, whereas the color palette learns the color distribution using color palettes. Vivid and diverse image colorization with a generative color prior (GCPrior) [11] is an image colorization method that learns color priors using a generative model. The color prior is the distribution of probable colors for each pixel in the image. The color prior is used by GCPrior to guide the colorization process, resulting in more vivid and diverse colorizations. DDColor [12] is an image-colorization approach that uses dual decoders. The pixel decoder reconstructs the spatial resolution of an image, whereas the query-based color decoder learns semantically aware color representations from multiscale visual data. The two decoders were merged using cross-attention to establish correlations between color and semantic information, substantially relieving the color-bleeding effect.

Transformers [13] have attracted significant interest in computer vision. Vaswani *et al.* [13] initially presented the transformer architecture. Subsequently, a novel approach to image classification was introduced, denoted as vision transformers (ViTs)[14]. In addition to image classification, transformers have been utilized in various other image-processing tasks, including object detection, segmentation, image super-resolution, denoising, and colorization. In addition, transformers, exemplified by ColTran[15], have exhibited encouraging outcomes in the image colorization task, thereby attesting to their efficacy in this domain. CT2 [16] is another example of image colorization that uses an end-to-end transformer framework. Grayscale features were extracted and encoded, and discrete color tokens representing quantized *ab* spaces were introduced. A dedicated color transformer fuses the image and color information guided by luminance selection and color attention modules.

Despite the notable performance improvement, existing colorization networks that rely on CNNs or transformers encounter notable limitations, including color bleeding, desaturation, and difficulties in capturing local and global features. We introduce a novel image colorization method to overcome these challenges that strategically integrates transformers and CNNs into the generator architecture.

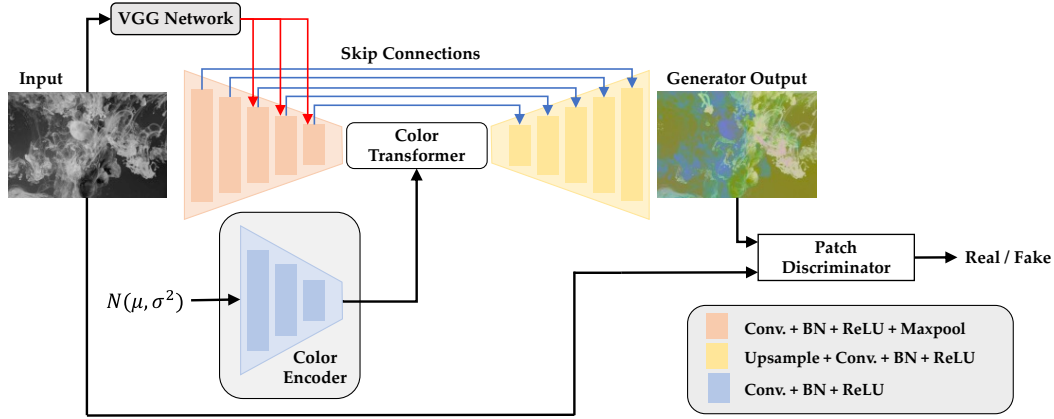

**Figure 1.** Overall architecture of the proposed network architecture.

This approach aims to effectively address the challenges of existing methods by leveraging the strengths of transformer architecture and adversarial training. Moreover, we introduce two key components in the generator, the color encoder and the color transformer, to further augment the colorization process. The color encoder focuses on capturing intricate color features, whereas the color transformer enhances the integration of local and global information using a transformer architecture. These modules form a comprehensive and robust image colorization framework, mitigating the shortcomings observed in the current state-of-the-art approaches.

## 3. Proposed Method

The proposed colorization method in this paper uses an encoder-decoder architecture with a color transformer at the bottleneck and a color encoder block in the generator. In this section, we describe the overall architecture of the proposed method. We then provide the details of the proposed generator architecture, which includes the color encoder, color transformer, and proposed objective function.

### 3.1. Overall Architecture

Figure 1 shows the overall architecture of the proposed image colorization network. The proposed method introduces a comprehensive architectural design that integrates several key components. Specifically, we employ VGG-based global feature extraction, a color encoder, a color transformer, and a GAN architecture to enhance the visual quality. Initially, the RGB color space is converted into the CIELAB color space (Lab) [17]. The laboratory separates luminance from chromaticity, thus providing a perceptually uniform space. *L* represents the luminance channel of an image, and *ab* represents the chrominance channels of the image. This separation helps the colorization model to capture chromatic details independent of luminance, thereby improving the overall accuracy and perceptual quality. The luminance channel image input undergoes initial processing via a pretrained VGG network and encoder, extracting high-level global features that capture semantic information. The global features from the VGG network are combined with the encoder layers, as shown in Figure 1. This integration of pretrained VGG features at different encoder levels is designed to enrich the understanding of the input image in the encoder, providing a more enhanced representation that facilitates improved colorization performance. Concurrently, a color encoder uses convolutional layers to produce color features from a normal distribution, as described in reference [2]. The integration of global and color-specific information is facilitated by fusing the color-encoded features at the bottleneck in the color transformer block and the global features in the encoder layers, as shown in Figure 1. The fused features are then fed into a Swin Transformer [18] block that captures the long-range dependencies and spatial relationships in the image. Two transformer blocks are used to capture the global information effectively. The decoder network employs a gradual upsampling process to reconstruct the ab channels of the lab color space while preserving fine-grained details using skip connections.

GAN architecture, which consists of a generator that includes an encoder, color transformer, color encoder, decoder, and discriminator, is utilized to improve visual fidelity. The generator tries to produce convincingly realistic colorizations, thereby deceiving the discriminator. In contrast, the discriminator's role is to differentiate between the colorized outputs and the actual color images that serve as the ground truth (GT). The training process is guided by various loss functions, such as perceptual loss, adversarial loss, and color loss, which collectively contribute to precise colorization. In our proposed architecture, we utilize a Patch-GAN-based discriminator [1] for image colorization. The Patch-GAN discriminator assesses local image patches instead of the entire image, allowing for a more detailed evaluation of textures and features. By concentrating on smaller regions, our method significantly enhances the synthesis of colorized images, achieving improved local coherence and a realistic distribution of textures, which contributes to the enhanced overall quality of the generated results.

### 3.2. Color Encoder

Color encoder plays a crucial role in the proposed image colorization by generating color features from a Gaussian normal distribution. This part utilizes a CNN to convert randomly sampled normal features into significant color-encoded features. Normal features are fed into the color encoder and subjected to multiple convolutional layers. These layers learn to extract spatially relevant information from the normal features, resulting in color-coded features that capture color-specific information.

To train the color encoder, the output of the VGG network is compared with the generated color-encoded features. The VGG network receives a color image input that comprises the *L* channel input image and the GT image *ab* channels. A VGG network can extract global features from a given color image, capturing high-level semantic information. The color-encoded features generated by the color encoder are compared with the global features extracted by the VGG network using $L_1$ loss. A color encoder is crucial for generating colorful and visually appealing colorization results.

### 3.3. Color Transformer

A color transformer module is designed to improve the image colorization process. This is achieved by integrating color features with grayscale image features and subsequently passing them through two Swin Transformer [18] layers, as shown in Figure 2. The fusion process generates a comprehensive representation by integrating global and color-specific information. Swin Transformer layers can capture global dependencies and spatial relationships, which facilitate the model's understanding of long-range dependencies and complex relationships present within the image. The incorporation of global information from a color transformer enhances the precision and visual quality of colorized outputs. A residual connection is used to compensate for missing information and improve the gradient flow. This process can be described as (1) – (5). First, outputs of encoder and color encoder $x_e$ and $x_{ce}$ are concatenated and single tensor $x_i$ is obtained as (1), setting the foundation for integrated feature processing.

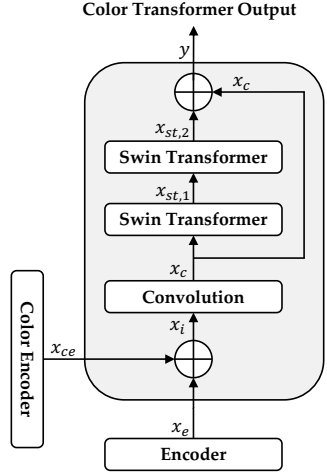

**Figure 2.** Color Transformer

$$x_i = Conc(x_e, x_{ce}), \quad (1)$$

Let $Conv(x_i)$ denotes a convolution operation that transforms $x_i$ by extracting spatial relationships. Then $x_c$ is obtained as (2).

$$x_c = Conv(x_i), \quad (2)$$

Subsequently, the $x_c$ is passed through two Swin Transformer blocks represented as $T_1^{sw}$ and $T_2^{sw}$. The Swin Transformer blocks enable the extraction and enhancement of long-range dependencies within the data.

$$x_{st,1} = T_1^{sw}(x_c), \quad (3)$$

$$x_{st,2} = T_2^{sw}(x_{st,1}), \quad (4)$$

Finaly, the obtained output $x_{st,2}$ is added elementwise to $x_c$ and represented as (5).

$$y = x_c + x_{st,2} \quad (5)$$

This addition of the initial convolutional features with the advanced features processed by the Swin Transformer blocks creates a residual connection that enhances the flow of gradients and compensates for any potential loss of information. where $y$ represents the output of color transformer module as in (5).

The color transformer ensures the seamless integration of grayscale and color features, facilitating a comprehensive understanding of the image content within the color transformer module. Incorporating Swin Transformers and the residual connection collectively enhances the capability of the model to produce accurate and visually compelling colorizations.

### 3.4. Objective Function

The objective function in the proposed method is defined as (6).

$$\mathcal{L} = \lambda_g \mathcal{L}_g + \lambda_p \mathcal{L}_p + \lambda_{L1}\mathcal{L}_1 + \lambda_c \mathcal{L}_c, \tag{6}$$

where $\mathcal{L}$ represents the total loss, $\mathcal{L}_g$ denotes the adversarial Wasserstein (WGAN) loss [19] and is used to avoid the vanishing gradient problem and achieve stable training for the GAN. $\mathcal{L}_p$ is the perceptual loss, which is the $L_2$ distance of the features extracted by the pretrained VGG network for the real and generated images.

A VGG loss function is used to improve the perceptual quality of the generated images. The VGG loss function is defined by the rectified linear unit activation layer of the pretrained VGG network.

$$L_p = \left\lVert \varphi_k(y) - \varphi_k(\tilde{y}) \right\rVert_2^2, \tag{7}$$

where $\varphi_k$ refers to the features of the k-th layer of the pretrained VGG network and $y, \tilde{y}$ represent the GT and output image, respectively.

$\mathcal{L}_{L1}$ in (6) is the conventional $L_1$ loss for the output colorized and GT images. $\mathcal{L}_c$ is the color loss, which is the comparison of the random normal distribution feature map from the color encoder and GT image feature map, and is defined as

$$\mathcal{L}_c = E \left\lVert G_f(N(\mu, \sigma^2)) - VGG(y) \right\rVert_1, \tag{8}$$

where, $N(\mu, \sigma^2)$ is the random normal distribution with mean $\mu = 0$ and standard deviation $\sigma^2 = 0.1$. $G_f$ represents the function of the color encoder, and $y$ is the GT image. $\lambda_g, \lambda_p, \lambda_{L1}$ and $\lambda_c$ values are fixed and empirically set to $\{\lambda_g, \lambda_p, \lambda_{L1}, \lambda_c\} = \{0.1, 100, 10, 1\}$, respectively.

### 4. Experimental Results

#### 4.1. Implementation Details and Results

The PASCALVOC [20] dataset, which contains 17,125 images, was used for training. A total of 15,413 images were used for training and 1,712 images were used for testing. The images were then rescaled to 256×256 pixels. The learning rate was set to $1 \times 10^{-4}$ and $2 \times 10^{-4}$ for the generator and discriminator, respectively. We use a batch size of 16 and an Adam optimizer with $\beta_1 = 0.5$ and $\beta_2 = 0.999$.

We used the peak signal-to-noise ratio (PSNR) [21], structural similarity index (SSIM) [22], and colorfulness [23] metrics to evaluate model performance. Colorfulness can be quantified mathematically by assessing the variation in color intensity within an image. The colorfulness is the standard deviation of the pixel values in the color channels. A higher standard deviation indicates greater color diversity and consequently, higher colorfulness. Colorfulness measures the difference in colorfulness values between the colored and GT images. Figrue 3 shows the colorization results of the proposed network.

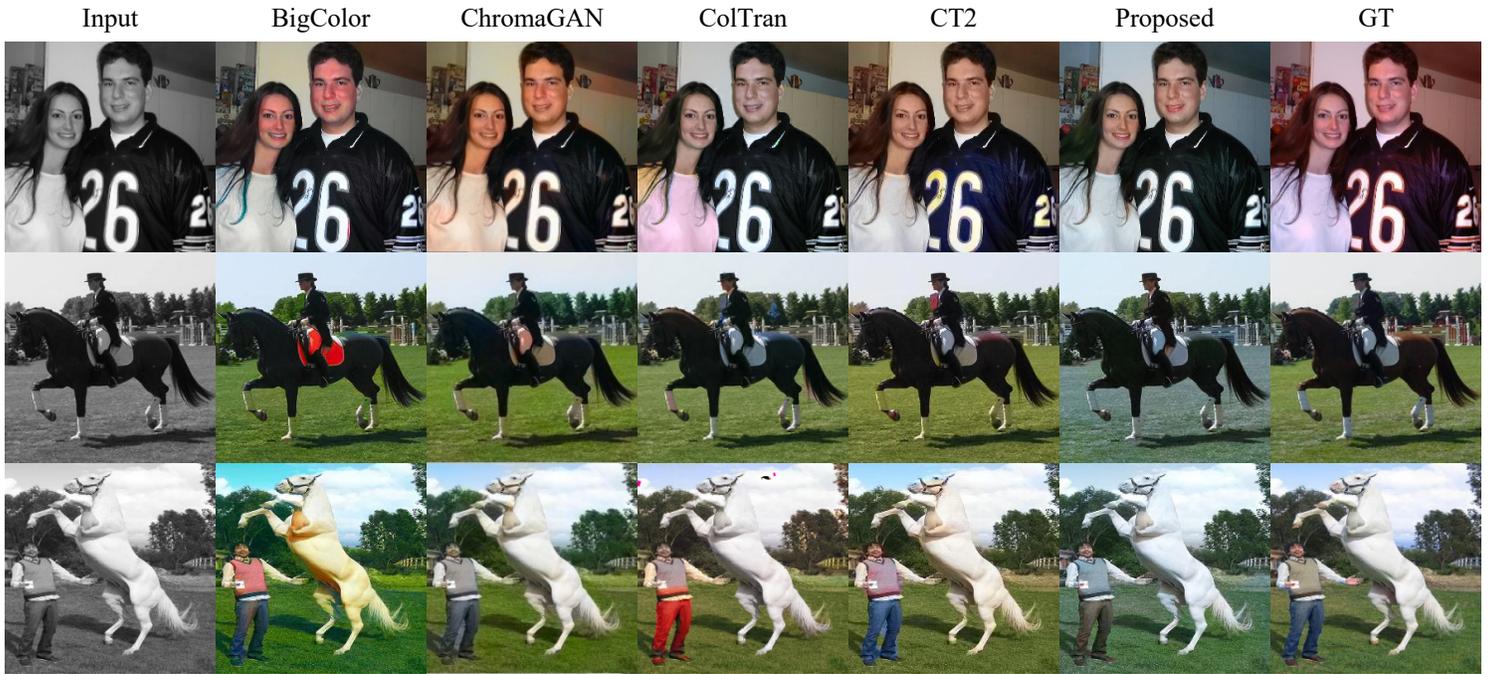

**Figure 3.** Qualitative comparisons.

Our method is compard with four fully automatic state-of-the-art image colorization networks: BigColor [8], ChromaGAN [21], ColTran [15], and CT2 [16]. Figure 3 shows that the proposed method yields more natural colors and outperforms the other methods. ChromaGAN [21] and ColTran [15] suffer from color bleeding artifacts and desaturation. BigColor and CT2 gave more vivid results but were far from GT. This is more evident in the 2nd row of Figure 3. More vividness results in unnatural colorization. As in the 3rd row of Figure 3, BigColor and CT2 give more vivid colors to a person in an image but lack realism and visual appeal. The proposed method output images that were more natural and closer to the GT colors.

We further evaluated our method using the PSNR, SSIM, and colorfulness. Our method outperformed the other methods in terms of the quantitative metrics, as shown in Table I. Higher PSNR and SSIM values indicated a more accurate and faithful representation of the colorized images than their GT counterparts. The ∆colorfulness values obtained in our experimental result indicate that our proposed method produces color variations that are close to GT images. Colorfulness has a higher value if rare colors are present in an image, regardless of color accuracy. The other methods may have higher colorfulness values, but the proposed method excels in ∆colorfulness, as the output image colors are closer to the GT. Figure 3 shows that the results of CT2 are more vivid but have bleeding artifacts. State-of-the-art methods show higher value in colorfulness due to more variations in color but are far from GT. The proposed network shows higher performance in ∆colorfulness values, which indicates that color is closely aligned to GT images.

TABLE I
QUANTITATIVE COMPARISONS IN PSNR (DB), SSIM, COLORFULLNESS AND ΔCOLORFULNESS

| Models | Evaluation metrics | | | |
|---|---|---|---|---|
| | PSNR(dB) | SSIM | Colorfulness | ΔColorfulness |
| BigColor [8] | 21.473 | 0.883 | 35.71 | 4.63 |
| ChromaGAN [21] | 23.636 | 0.882 | 21.89 | 9.19 |
| ColTran [15] | 23.839 | 0.868 | 35.74 | 4.66 |
| CT2 [16] | 19.304 | 0.912 | **36.04** | 4.96 |
| Proposed | **24.023** | **0.941** | 27.22 | **3.86** |

TABLE II
RESULTS OF ABLATION STUDIES

| Models | Evaluation metrics | | | |
|---|---|---|---|---|
| | PSNR(dB) | SSIM | Colorfulness | ΔColorfulness |
| A (UNet) | 23.11 | 0.936 | 13.97 | 17.11 |
| B (w/o Color Encoder) | 23.48 | 0.937 | 19.15 | 11.93 |
| C (w/0 Color Transformer) | 22.97 | 0.937 | 23.29 | 7.79 |
| Proposed | **24.02** | **0.941** | **27.22** | **3.86** |

## 4.2. Ablation Studies

Ablation studies were performed to evaluate the individual contributions of the color transformer and color encoder components in the proposed method.

- A (U-Net): The baseline UNet model was developed by removing the color encoder and transformer from the architecture. UNet provides standard performance across all metrics.

- B (without Color Encoder): We removed the color encoder and evaluated the performance of the color transformer using the proposed method. Excluding the color encoder resulted in improved PSNR and colorfulness, with a reduction in Δcolorfulness as shown in Table II.

- C (without a Color Transformer): We removed the color transformer and evaluated the performance of the color encoder in the proposed network. Removing the color transformer decreased the PSNR. The colorfulness value is increased, accompanied by a notable reduction in Δcolorfulness. A notable reduction in Δcolorfulness indicated a decrease in perceptual differences when compared to the baseline.

- Proposed: The proposed method achieved the highest PSNR and SSIM, indicating superior image colorization quality. Notably, colorfulness reached its peak, and Δcolorfulness was minimized, indicating the effectiveness of both color encoder and color transformer in enhancing colorization performance.

## 5. Conclusion

This study concludes that the integration of a color transformer, color encoder, and GAN results in a novel image colorization method that demonstrates significant progress in image colorization. Our approach utilizes a transformer architecture to gather global information efficiently and incorporates the realistic colorization capabilities of GANs to produce precise and visually appealing colorization results. The fusion of color features from the color encoder with grayscale image features enhances the overall colorization process, preserves fine-grained details, and produces a high-quality output. Experimental evaluations demonstrate the superiority of our approach over other state-of-the-art methods. This research makes a valuable contribution to the progress of image colorization methods, thereby opening up possibilities for their use in the digital restoration, entertainment, and analysis of historical images. Future work may explore further optimizations and extensions to address specific challenges and continue to enhance the accuracy and realism of the colorization outputs.